\documentclass[lettersize,journal]{IEEEtran}
\usepackage{amsmath}
\usepackage{graphicx} 
\usepackage{algorithm}
\usepackage{algpseudocode}
\usepackage{mathtools}
\usepackage{dsfont}
\usepackage{chemformula}
\usepackage{esvect}
\usepackage{amsmath,bm}

\title{Autonomous Vision-Based Magnetic Microrobotic Pushing of Micro-Objects and Cells}

\author{Max Sokolich$^{1}$, Ceren Kirmizitas$^{1}$, Sambeeta Das$^{1}$, Ron Weiss$^{2}$ 

\thanks{$^{1}$Department of Mechanical Engineering, University of Delaware, Newark, DE 29716, USA
        {\tt\footnotesize \{sokolich,ceren,samdas\}@udel.edu}}%

\thanks{$^{2}$Department of Biological Engineering, Massachusetts Institute of Technology, Cambridge, MA 02142, USA
        {\tt\footnotesize rweiss@mit.edu}}%
        
}

\date{\today}

\begin{document}
\maketitle

\begin{abstract}
Accurate and autonomous transportation of micro-objects and biological cells can enable significant advances in a wide variety of research disciplines. Here, we present a novel, vision-based, model-free microrobotic pushing algorithm for the autonomous manipulation of micro objects and biological cells.  The algorithm adjusts the axis of a rotating magnetic field that in turn controls the heading angle and spin axis of a spherical Janus rolling microrobot. We introduce the concept of a \textit{microrobotic guiding corridor} to constrain the object and to avoid pushing failures. We then show that employing only two simple conditions, the microrobot is able to successfully and autonomously push microscale objects along predefined trajectories. We evaluate the performance of the algorithm by measuring the mean absolute error and completion time relative to a desired path at different actuation frequencies and guiding corridor widths. Finally, we demonstrate biomedical applicability by autonomously transporting a single biological cell, highlighting the methods potential for applications in tissue engineering, drug delivery and synthetic biology. 
\end{abstract}

\section{Introduction}
Micro-robotics \textemdash{} that is, the application of macroscale robotic principles such as automation, actuation, and sensing to the microscale \textemdash{} has attracted significant interest in recent decades \cite{nelson2010microrobots}. Applications ranging from targeted drug delivery \cite{jang2019targeted} to environmental remediation \cite{jancik2025nano} have shown notable promise in using microrobots to address pressing real-world challenges. Microrobots are typically actuated using magnetic fields \cite{xu2015magnetic}, light \cite{bunea2021light}, acoustic fields \cite{aghakhani2020acoustically}, chemical gradients \cite{truby2023chemically}, thermal gradients \cite{erdem2010thermally} or by harnessing innate biological propulsion mechanisms \cite{alapan2019microrobotics}. These unique actuation mechanisms arise from difficulty incorporating macro-scale actuators, sensors, power supplies and control logic into entities often smaller than the width of a human hair. 

Of the many potential applications of microrobots, microscale manipulation holds significant promise especially in areas of biology \cite{jager2000microrobots}. Manipulating single cells in a potentially non-invasive, wireless manner can enable breakthroughs in stem cell therapy \cite{jeon2021magnetically}, personalized medicine \cite{garcia2021role}, and artificial organoid manufacturing \cite{mohanty2025magnetized}. To fully realize the potential of micro-robotics in biology, new control algorithms, control systems, and fabrication techniques must be developed. This work aims to advance the field of microscale manipulation via a novel autonomous pushing algorithm using rolling magnetic microrobots. 

To date, there have been numerous approaches to both wired and wireless microscale manipulation \cite{savia2009contact}. For wired approaches, \cite{lynch2008vision} created a vision based micro-particle pushing scheme using an atomic microscopy probe able to manipulate spheres on the order of 5\,microns. Similarly, \cite{10.1117/12.325735} also used an AFM probe for micro-object pushing.  In \cite{shahini2013automated}, researchers designed and experimentally validated a controller for pushing micro objects in the workspace using a tethered cantilever pushing mechanism.  Moreover, \cite{5342467} developed a wired robotic system with a micro-gripper for picking and placing microspheres in the workspace. 

\begin{figure}[h! tbp]
    \centering
    \includegraphics[width=\linewidth]{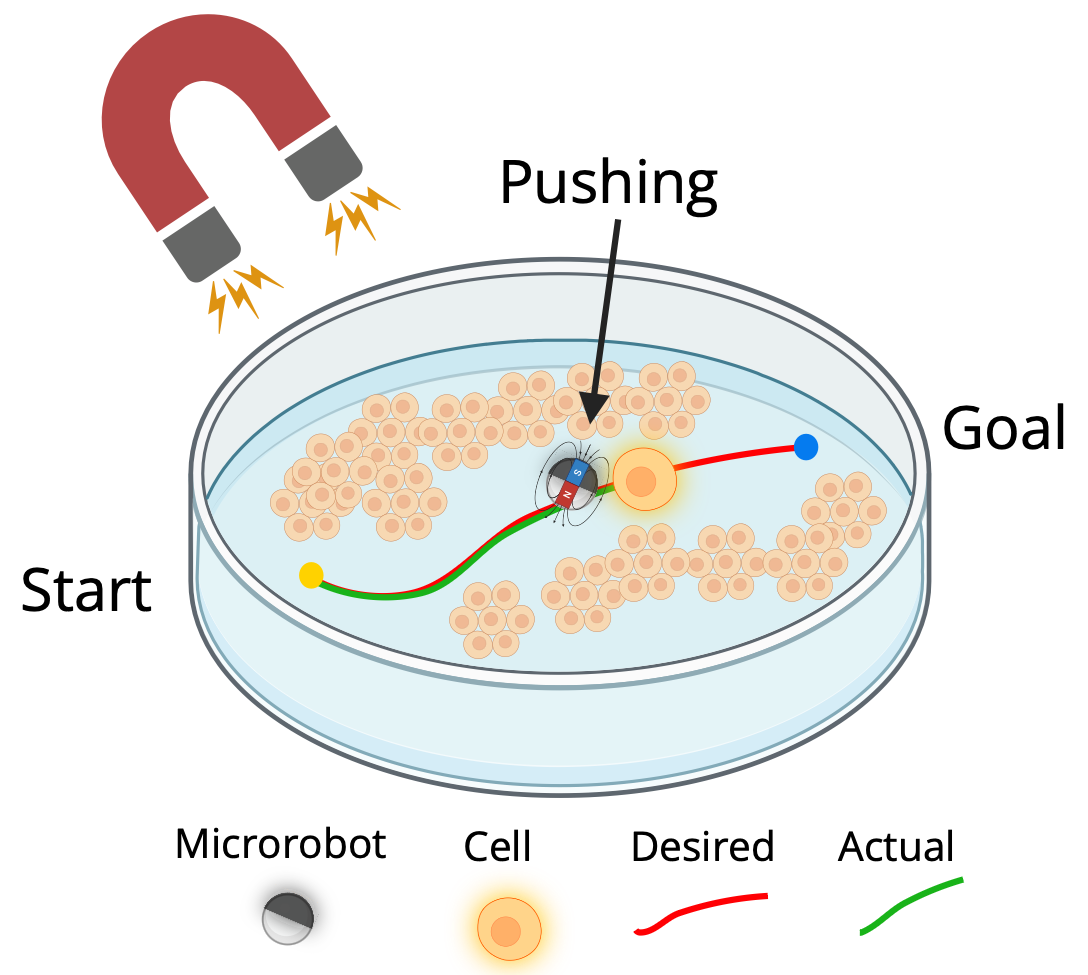}
    \caption{Graphical Abstract}
    \label{fig:1}
\end{figure}

For approaches using wireless magnetic gradients, \cite{8376001} developed a MEMS-based micro-force sensing microrobot approximately 4\,$\mu$ms in size. Manipulation forces were measured by calculating the deformation of the end effector via optical feedback. In \cite{6085616}, researchers used an approximately 400\,$\mu$m Mag-$\mu$bot with magnetic gradients to push approximately 210 $\mu$m in size passive microspheres in 2D. They also developed autonomous control policies. Additionally, \cite{steager2013automated} demonstrated the automatic transport of micro-beads and single cells using a magnetic gradient-based micro-pusher, along with an autonomous control procedure for delivering micro-beads to single cells. Hagiwara et al. demonstrated a lab-on-chip microrobotic device for single-cell positioning and control using a permanent magnet \cite{hagiwara2011chip}. Finally, authors in \cite{duygu2025real} introduced a haptic 3D magnetic tweezer system capable of transporting micro-objects, navigating microfluidic channels, and even implemented a novel disassembly method.  

For wireless approaches using rotating magnetic fields, researchers have controlled swarms of \ch{Fe3O4} nanoparticles to engulf passive spheres and transport them to user-defined positions \cite{10598230}. In \cite{8268550}, a system was demonstrated that enabled both rolling and more complex kayaking motions for the manipulation and release of micro-objects using a rotating magnetic field. A peanut-shaped microrobot approximately 3,$\mu$m in size was shown to transport microparticles using open-loop control. Additionally, \cite{5229326} demonstrated the micromanipulation of passive microspheres using an untethered, electromagnetically actuated magnetic microrobot, capable of both contact and non-contact pushing modes. In \cite{zhang2012targeted}, the authors used nickel-coated nanowires driven by a rotating magnetic field to selectively push a variety of microscale objects, including individual flagellated microorganisms and human blood cells. Moreover, \cite{zhou2017dumbbell} introduced a dumbbell-shaped magnetic microrobot capable of on-demand transport of micro-objects via tumbling, wobbling, and rolling motions. Finally, in previous work, we have demonstrated autonomous control and cellular obstacle avoidance using magnetized cell-bots \cite{sokolich2024janus} as well as open loop manipulation of cells using rolling microrobots \cite{rivas2022cellular}.

Finally, non-magnetic actuation methods have also been used for microscale manipulation. Ahmed et al. described an acoustic-based, lab-on-chip manipulation technique capable of rotating single microparticles, cells, and organisms \cite{ahmed2016rotational}. Additionally, holographic optical tweezers have been used to transport various cell types of different sizes \cite{hu2016automated}. Buican et al.  demonstrated single-cell manipulation using optical tweezer light trapping \cite{buican1987automated}. Chemically actuated micro-jets have also been shown to transport multiple cells to specific locations in a fluid, despite the toxic effects of the fuel \cite{sanchez2011controlled}. Palacci et al. introduced a self-propelled colloidal hematite "docker" that can steer toward, dock with, and transport colloidal cargo many times its own size \cite{palacci2013photoactivated}. The self-propulsion and docking processes are reversible and activated by visible light. Additional work that inspired this algorithm include  \cite{krivic2019pushing}, \cite{7030011}, \cite{10610098}, \cite{6878041}, \cite{ye2012micro}, and \cite{khalil2020controlled}.  

The contributions of this article include a novel,  vision-based, model-free, data-driven, geometric algorithm for autonomously pushing a micro-object using a spherical magnetic rolling microrobot. We demonstrate the algorithm in experiment and compare the performance of the algorithm by measuring the completion time following a predefined path, and the mean absolute error between actual and desired path. Finally, we showcase biomedical applicability of the algorithm through the manipulation of a single biological cell.

The remainder of this article is organized as follows: We briefly formulate the problem in section II and describe the experimental setup in section III. We present the algorithms control methodology in section IV, followed by a description of the performance metric in section V. We experimentally validate and evaluate the performance of the algorithm in section VI. Finally, we end with a brief discussion on future work to improve the algorithm in section VII, and a conclusion in section VIII.

\section{Problem Formulation}
The current work addresses the problem of wireless microscale manipulation using magnetic microrobots. The work aims to design a model-free, data-driven algorithm for wireless and autonomous transportation of microscale objects.  The goal is to autonomously transport the object accurately along a user-defined trajectory (see Figure \ref{fig:1}). 

\begin{figure}[h!]
    \centering
    \includegraphics[width=\linewidth]{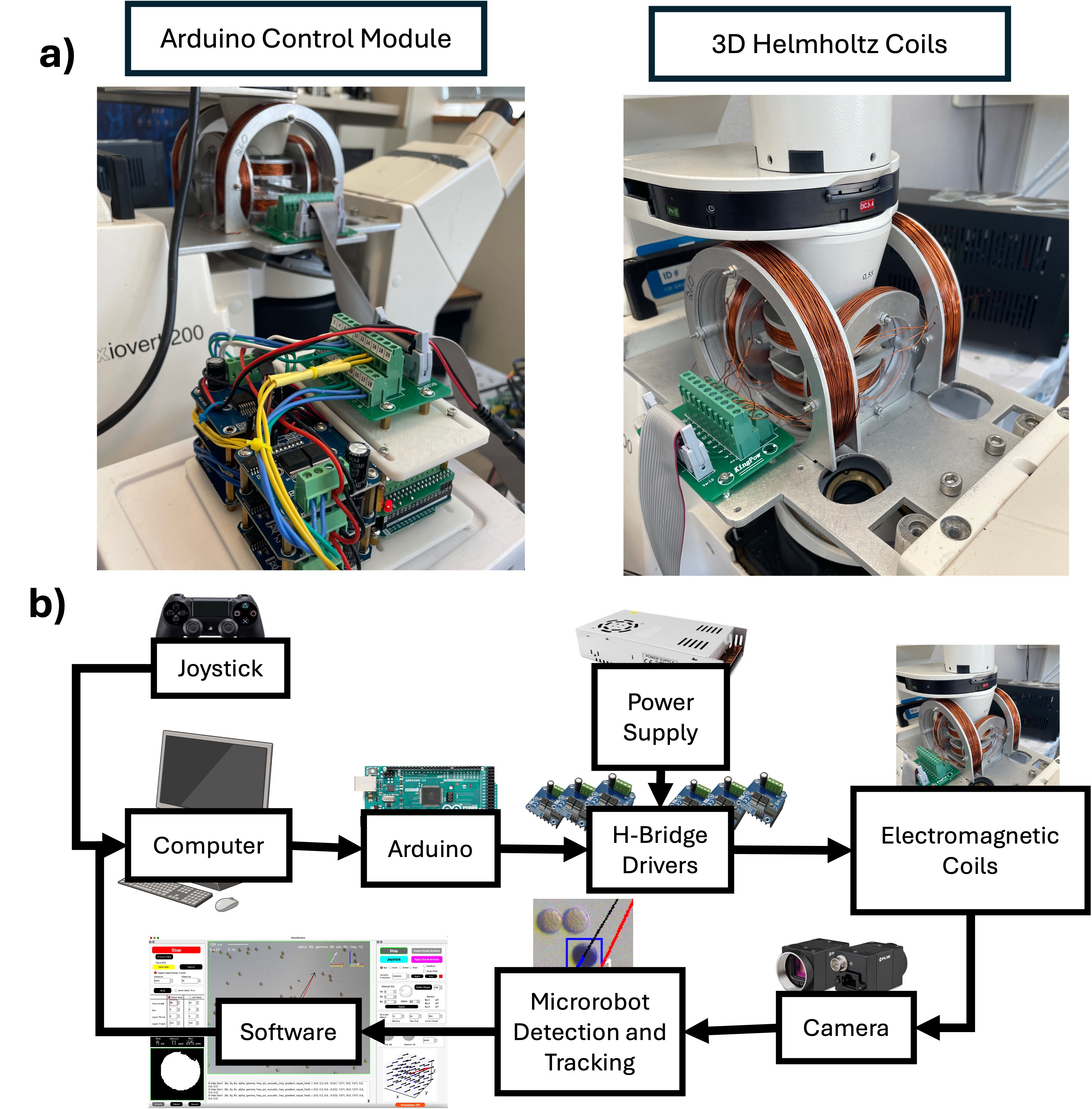}
    \caption{Experimental Setup. a) Arduino control module and 3D Helmholtz coil system mounted on a Zeiss Axiovert 200 inverted microscope. b) System information flowchart.}
    \label{fig:2}
\end{figure}

\section{Experimental Setup}
\begin{figure}[h!]
    \centering
    \includegraphics[width=\linewidth]{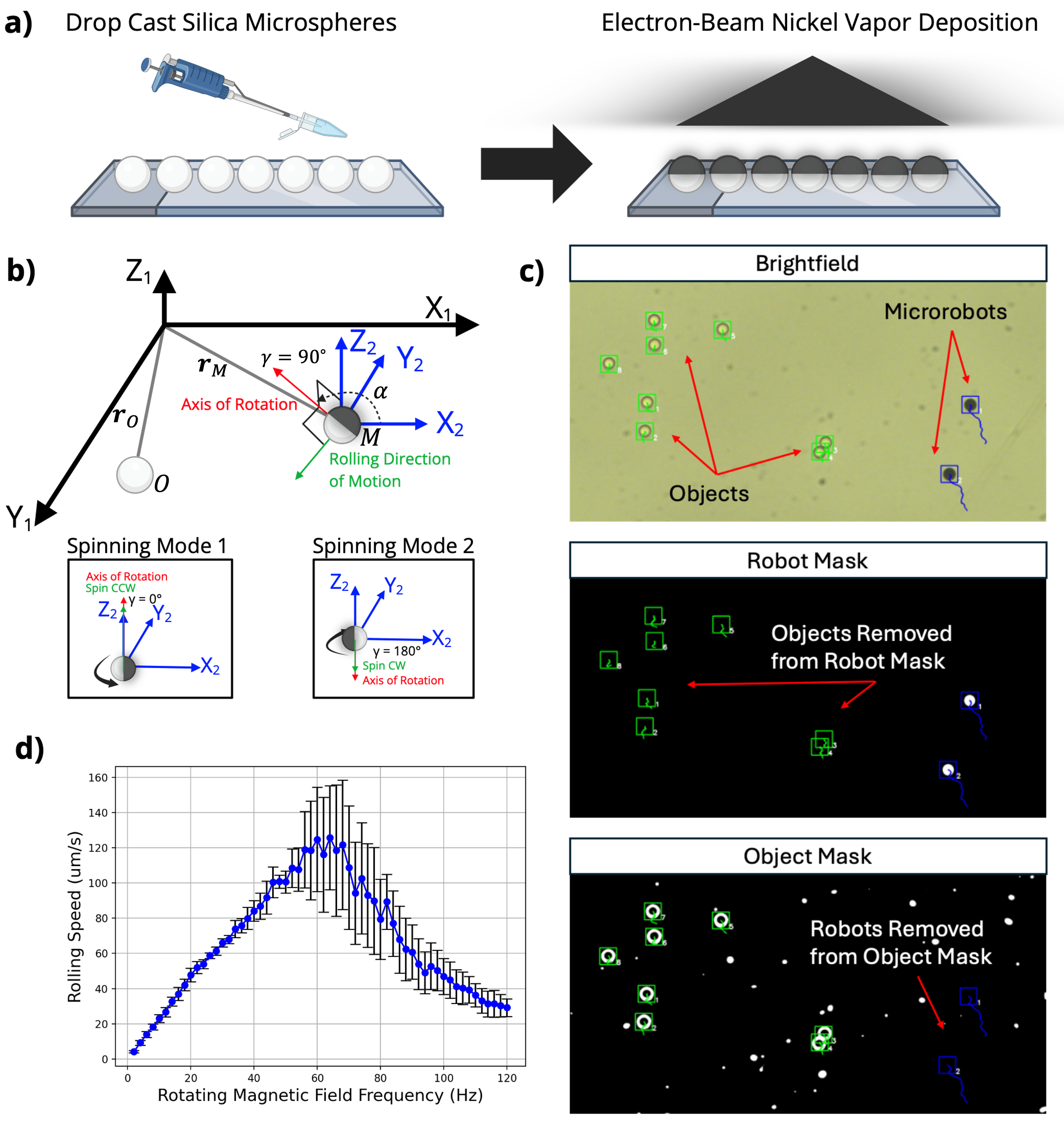}
    \caption{a) Procedure for fabricating magnetic rolling microrobots using electron beam vapor deposition. b) Image and positional tracking coordinate system in black. Actuation coordinate system in blue. Heading angle $\alpha$, attitude angle $\gamma$ and resulting axis of rotation labeled for different actuation modes. c) Brightfield, robot mask and cell mask screenshots.  d) Speed vs frequency graph of 10\,$\mu$m magnetic rolling microrobot in DI on a glass microscope slide. The step-out frequency was measured to be approximately 60Hz. }
    \label{fig:3}
\end{figure}

The microrobots are fabricated by first plasma-cleaning a glass microscope slide on high power for 5 minutes using a Harrick Plasma PDC-001 Plasma Cleaner. 10\,$\mu$m  diameter silica microspheres (Alpha Nanotech Batch $\#$01712) are then drop-cast onto the surface of the slide. An electron beam vapor deposition instrument is used to coat the slide with 100\,nm of nickel, rendering the microrobots magnetic (see Figure \ref{fig:3}a). The microrobots can then be scraped off the surface using a pipette and transferred to a separate microscope slide for further experimentation.

A six coil, 3D Helmholtz electromagnetic manipulation system was used to actuate the magnetic microrobots (See figure \ref{fig:2}). The system consists of 3 pairs of Helmholtz coils arranged in the X, Y and Z directions and can be described in more detail in \cite{sokolich2023modmag}. A uniform field can be generated along each axis. The coils are mounted on a Zeiss Axiovert 200 inverted microscope and a 20x objective is used. A FLIR BFS-U3-50S5C-C USB 3.1 Blackfly® S Color Microscope Camera is used to record live images from the workspace. The FPS of the camera is set to 24 fps.

An Arduino Mega 2560 is used to output the time varying rotating magnetic field signals needed to actuate a magnetic rolling microrobot. However, the current output from the Arduino is too weak to adequately power the electromagnets, therefore the current must be amplified. This is done using six SEEU. AGAIN BTS7960B 43A Double DC Stepper H-Bridge PWM drivers connected to each of the six coils.  These motor drivers use the low power PWM signal from the Arduino to switch the higher power circuit between an external power supply and the electromagnets. As a result, we can control the strength and polarity of each electromagnet by varying the duty cycle and direction of the PWM signal. At 100$\%$ duty cycle, the field was measured with a Teslameter to be 3\,mT, 5\,mT, and 13\,mT in the x,y, and z directions respectively. 

We use the following equations to control the axis of the rotating magnetic field, 
\begin{equation} \label{eq:magnet}
\begin{split}
    \mathbf{B} =
    \begin{bmatrix}
        -\cos(\gamma)\cos(\alpha)\cos(-2\pi f t) - \sin(\alpha)\sin(-2\pi f t) \\
        -\cos(\gamma)\sin(\alpha)\cos(-2\pi f t) + \cos(\alpha)\sin(-2\pi f t) \\
        \sin(\gamma)\cos(-2\pi f t)
    \end{bmatrix},
\end{split}
\end{equation}
where $\gamma\in[0, 180^{\circ}]$ is the elevation angle, $\alpha\in[0, 360^{\circ}]$ is the azimuthal angle, and $f\in\mathds{R}$ is the frequency of rotation. $f$ is also equal to $\omega /2\pi$ where $\omega$ is the angular velocity. By varying $\gamma$ and $\alpha$, we can effectively adjust the axis of the rotating magnetic field within the Helmholtz workspace. To control the heading angle of a rolling microrobot along the XY plane, we add $\pi/2$ to $\alpha$, and set $\gamma=90^{\circ}$. Additionally, setting $\gamma$ to $0^{\circ}$ spins the microrobot in the counterclockwise direction (CCW), and clockwise (CW) when $\gamma$ is $180^{\circ}$. See Figure \ref{fig:3}b for a coordinate system representation. It should be noted that a negative $\omega$ implies counterclockwise rotation and a positive $\omega$ indicates a clockwise rotation. To maintain right handed chirality of the field, the angular frequency $\omega = 2 \pi f$ is negative in equation \ref{eq:magnet}. 

The frequency of rotation $f$ controls the speed of the microrobot and theoretically follows $v=\omega r= 2 \pi f r$ where $\omega$ is the angular velocity. As $f$ increases, so does the speed of the rolling microrobot up until a certain step-out frequency. This is depicted in Figure \ref{fig:3}d. The step-out frequency is the maximum frequency of the rotating magnetic field for which the robot can synchronously align with the field without slipping or losing synchronization.  At high frequency, the magnetic torque is no longer sufficient to overcome hydrodynamic drag and frictional resistance, leading to de-synchronization resulting in a decrease in speed. The step-out frequency was found to be approximately 60\,Hz.

The actuation variables $\alpha$, $\gamma$, and $f$ can be adjusted in real time using custom control software written in Python's PyQt5 library. The commands are then sent from the control software to the Arduino Mega over Serial Communication protocol.

The control software also integrates real-time camera feedback to track and detect objects in the workspace using OpenCV python. It is important to note that the coordinate system in OpenCV defines the origin at the top left of the image.  The Spinnaker SDK, or more specifically, the PySpin API and EasyPySpin python library allow straightforward access to the raw camera frame using python.  The software allows the user to select a microrobot, an object to be manipulated, and a target coordinate or trajectory.  The tracking process begins by converting the incoming color frame from the microscope camera into a binary black and white mask. Users can adjust parameters such as the upper and lower threshold values, blur radius, and dilation kernel size within the software to optimize the mask and maximize contrast between the microrobot and the background. Once the mask parameters are configured, the user can click near the microrobot in the imaging interface, prompting the software to calculate a bounding box around the microrobot. The center of mass within this bounding box is then calculated, allowing the program to determine the position of the microrobot within the workspace.  As the microrobot moves, its position shifts within the bounding box. The software continuously recalculates the center of mass inside the bounding box and adjusts the box to track the new position of the microrobot in real-time.

To differentiate between the microrobot and the object, separate masks are calculated simultaneously. This allows the user to adjust each mask independently, which is particularly useful when the microrobot and object differ in size, shape, or color. Since the microrobot and object are often in close proximity during manipulation, additional functions are implemented in the software to prevent the trackers from merging. Specifically, the microrobot is removed from the object's mask, and the object is removed from the microrobot's mask. See figure \ref{fig:3}c for illustrations of the masking process.

\section{Control Theory}
\begin{figure*}[h! tbp]
    \centering
    \includegraphics[width=0.97\textwidth]{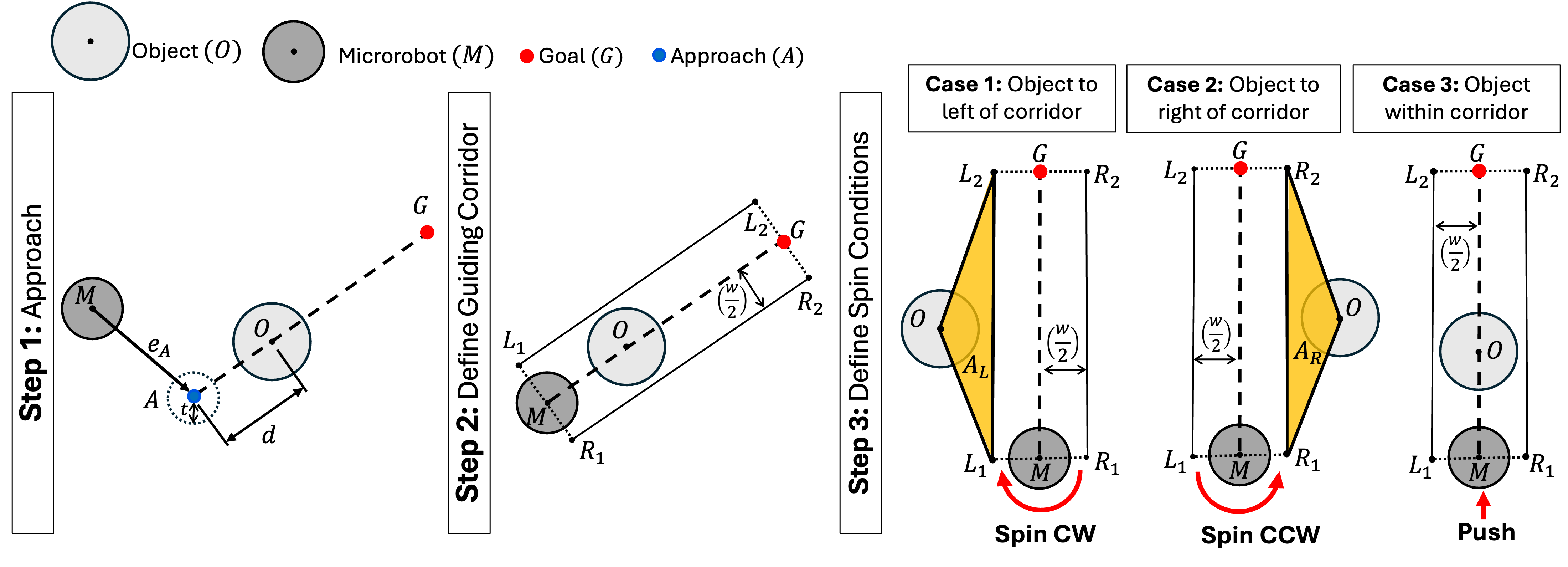}
    \caption{Schematic of pushing algorithm}
    \label{fig:4}
\end{figure*}

In overview, the algorithm works by controlling the axis of the rotating magnetic field, which determines the spinning axis of the magnetic microrobot (see Figure \ref{fig:3}b). By dynamically adjusting this axis, the system can manipulate the microrobot’s orientation and position. The algorithm can be described in three steps: the approach, the guiding corridor, and the spin conditions, as shown in Figure \ref{fig:4}. For clarity, the following notation is used throughout to describe the algorithm. The position vector of a microrobot at point $M$ with coordinate $(x_M, y_M)$ is denoted $\mathbf{r}_M = [x_M, y_M]^T$. The position vector of the object at point $O$ with coordinate $(x_O, y_O)$ is denoted $\mathbf{r}_O = [x_O, y_O]^T$. Finally, the position vector of the goal at point $G$ with coordinate $(x_G, y_G)$ is denoted $\mathbf{r}_G = [x_G, y_G]^T$. The algorithm begins once a microrobot, a desired object for manipulation, and a target coordinate or trajectory have all been selected and defined in the software.

\subsection{Approach}
In step 1 of the algorithm, the microrobot is guided to an approach coordinate $A = (x_A, y_A)$. See Figure \ref{fig:4} This coordinate is determined by calculating the unit vector from $\mathbf{r}_O$ to $\mathbf{r}_G$ and then subtracting a user defined distance $d$ from $\hat{\mathbf{r}}_{OG}$.  The microrobot is then actuated toward the approach coordinate by setting the heading angle to
\begin{equation}
\alpha = \arctan{\left(\frac{-(y_A - y_M)}{x_A - x_M}\right)}.
\end{equation}
It should be noted that the numerator must be negative to compensate for the y-direction difference between the image tracking coordinate system and the actuation coordinate system. The error between the microrobot $M$ and the approach coordinate $A$ is continuously calculated using  $e_A = \sqrt{(x_A - x_M)^2 + (y_A-y_M)^2}$. The algorithm concludes the microrobot has arrived at the approach coordinate once $e_A$ is less than a user defined threshold $t$.

\subsection{Guiding Corridor}
Once the microrobot has arrived at the approach coordinate, a guiding corridor is calculated beginning at the robots current position $M$ and ending at the goal position $G$. This is done by finding the coordinates $L_1, L_2, R_1, R_2$ as seen in Figure \ref{fig:4}. These are calculated by first defining a vector that extends from $M$ to $G$ using 
\begin{equation}
\mathbf{r}_{MG} = \mathbf{r}_{G} - \mathbf{r}_{M}=
\begin{bmatrix}
x_G - x_M \\
y_G - y_M
\end{bmatrix}. 
\end{equation}

Then, a unit vector perpendicular to $\mathbf{r}_{MG}$ is calculated using,
\begin{equation}\label{unit}
\hat{\mathbf{r}}_{MG}^\perp = \frac{1}{\|\mathbf{r}_{MG}\|} 
\begin{bmatrix}
-(y_G - y_M) \\
x_G - x_M
\end{bmatrix}
\end{equation}

Finally, the coordinates $L_1, L_2, R_1, R_2$ that define the guiding corridor with a width $w$ can be calculated using,

\begin{equation}\label{L1}
L_1 = (x_{L_1}, y_{L_1}) =
(x_M, y_M) + \frac{w}{2}\hat{\mathbf{r}}_{MG}^\perp
\end{equation}

\begin{equation}\label{L2}
L_2 = (x_{L_2} ,y_{L_2}) = (x_G, y_G) + \frac{w}{2}\hat{\mathbf{r}}_{MG}^\perp
\end{equation}

\begin{equation}\label{R1}
R_1 = (x_{R_1}, y_{R_1}) = (x_M, y_M) - \frac{w}{2}\hat{\mathbf{r}}_{MG}^\perp
\end{equation}

\begin{equation}\label{R2}
R_2 = (x_{R_2},y_{R_2}) = (x_G, y_G) - \frac{w}{2}\hat{\mathbf{r}}_{MG}^\perp
\end{equation}

\subsection{Spin Conditions}
Once the guiding corridor is defined, the algorithm will attempt to push the object towards the goal coordinate. This is depicted in case 3 of Figure 4. This is done by setting the heading angle to
\begin{equation}
\alpha = \arctan{\left(\frac{-(y_G - y_M)}{x_G - x_M}\right)}.
\end{equation}
To detect whether the object has deviated from the guiding corridor and needs adjustment, two conditions (or cases) are employed, as shown in Figure \ref{fig:4}. Both cases follow the same logic to determine if the object has moved outside the guiding corridor.

In case 1, the object is to the left of the guiding corridor and needs to be repositioned. This is done by finding the sign of the triangular area formed by the coordinates $L_1$, $L_2$, and $O$ which is shaded in orange in figure \ref{fig:4}. In other words, the relative position of the vector $\mathbf{r}_{{L_1}{O}}$ with respect to the vector $\mathbf{r}_{{L_1}{L_2}}$ can be determined using the sign of this area. This area is also equivalent to the magnitude of the cross product of the vectors $\mathbf{r}_{{L_1}{L_2}}$ and $\mathbf{r}_{{L_1}{O}}$ can be calculated as follows,

\begin{equation}\label{ALdet}
    A_L = \frac{1}{2}\lvert\lvert \mathbf{r}_{{L_1}{L_2}} \times \mathbf{r}_{{L_1}{O}}\rvert\rvert = 
    \frac{1}{2} \begin{vmatrix}
     x_{L_1} & y_{L_1} & 1 \\
     x_{L_2} & y_{L_2} & 1 \\
     x_{O} & y_{O} & 1 \\
    \end{vmatrix}.
\end{equation}

\begin{equation}\label{AL}    
\begin{split}
    A_L = \frac{1}{2}x_{L_1}(y_{L_2}-y_O)-y_{L_1}(x_{L_2}-x_O)
    \\+x_{L_2}y_O-y_{L_2}x_O.
\end{split}
\end{equation}

If $A_L < 0$, this indicates that the object is to the left of the vector $\mathbf{r}_{{L_1}{L_2}}$.  As a result, the algorithm will reposition the object by spinning the microrobot in the clockwise direction. This is done by setting $\gamma = 180^{\circ}$. This induces a microfluidic vortex that attempts to manipulate the object CW back into the guiding corridor.

Similarly, in case 2, if the object is to the right of the guiding corridor, the algorithm must also reposition the object. This is detected by finding the sign of the area $A_R$ of the triangle formed by $R_1$, $R_2$, and $O$ which is also shaded in yellow in figure \ref{fig:4}. The signed area $A_R$ is calculated in similar way to $A_L$ as follows,

\begin{equation}\label{ARdet}
    A_R = \frac{1}{2}\lvert\lvert \mathbf{r}_{{R_1}{R_2}} \times \mathbf{r}_{{R_1}{O}}\rvert\rvert = 
    \frac{1}{2} \begin{vmatrix}
     x_{R_1} & y_{R_1} & 1 \\
     x_{R_2} & y_{R_2} & 1 \\
     x_{O} & y_{O} & 1 \\
    \end{vmatrix}.
\end{equation}

\begin{equation}\label{AR}    
\begin{split}
    A_R = \frac{1}{2}x_{R_1}(y_{R_2}-y_O)-y_{R_1}(x_{R_2}-x_O)
    \\+x_{R_2}y_O-y_{R_2}x_O.
\end{split}
\end{equation}

In this case, if $A_R > 0$, this indicates that the object is to the right of the vector $\mathbf{r}_{{R_1}{R_2}}$ and therefore must be repositioned in the counterclockwise direction. This is done by setting $\gamma = 0^{\circ}$. 

These 3 cases allow the microrobot to adequately push an object towards a goal, and reposition the object back into the guiding corridor when needed.  To determine if the object has arrived at the target position, the error between the object $O$ and the target $G$ is continuously calculated. The error is calculated using $e_G = \sqrt{(x_G - x_O)^2+ (y_G-y_O)^2}$. If this error is less than a user-defined threshold value $t$, we conclude the object has arrived at the destination, and the algorithm will terminate.  By linking together multiple goal coordinates in a list, a nonlinear path can be generated that allows the microrobot to push objects along more sophisticated trajectories. Instead of the algorithm terminating once it reaches a single goal coordinate, the algorithm will move on to the next goal coordinate. 

\begin{algorithm}
 \caption{Microrobotic Micro-vortice Readjustment Pushing Algorithm}
 \begin{flushleft}
        \textbf{Input:} {Object Desired Goal Position $G =(x_G, y_G)$}\\
        \textbf{Output:} {Magnetic field vector $\textbf{B}= [B_x,B_y, B_z]^T$}
\end{flushleft}
\begin{algorithmic}[1]
\State{Move to approach coordinate $A \gets (x_A, y_A)$}
\State{Define $\alpha, \gamma, f$}
\While{$e_G > t$}
\State{$M \gets (x_M, y_M)$}
\State{$O \gets (x_O, y_O)$}
\State{calculate $L_1, L_2, R_1, R_2$}
\State{calculate $A_R$, $A_L$}
\State{calculate $e_G  \leftarrow \sqrt{x_G - x_O)^2+ (y_G-y_O)^2}$ }
\If{$D_L < 0$}
\State{Spin CW}
\State{$\gamma \gets 180^{\circ}$}
\ElsIf{$D_R > 0$}
\State{Spin CCW}
\State{$\gamma \gets 0^{\circ}$}
\Else
\State{Push}
\State{$\gamma \gets 90^{\circ}$}
\State{$\alpha  \leftarrow \arctan{\bigg(\dfrac{ -(y_G-y_M)}{(x_G - x_M)}\bigg)}$}
\EndIf
\small
    \State{$B_x \gets -\cos{(\gamma)} \cos{(\alpha)} \cos{(-2 \pi f t)} - \sin{(\alpha)} \sin{(-2 \pi f t)}$}
    \State{$B_y \gets -\cos{(\gamma)} \sin{(\alpha)} \cos{(-2 \pi f t)} + \cos{(\alpha)} \sin{(-2 \pi f t)}$}
    \State{$B_z \gets \sin{(\gamma)} \cos{(-2 \pi f t)}$}
    \normalsize
\EndWhile 
\end{algorithmic}
\label{alg:alg1}
\end{algorithm}
\normalsize

In summary, the algorithm begins by guiding the microrobot to an approach coordinate $A$, which is in line with the object $O$ and the goal position $G$. Once arrived at the approach coordinate, the microrobot attempts to push the object toward the goal. The tracking error is defined as the Euclidean distance between the object’s current position and the goal. If the error is less than a predefined threshold value $t$, the algorithm concludes that the object has reached the goal. If a desired trajectory consisting of $N_d$ desired goal positions is defined, the algorithm proceeds to the next goal in the list instead. This enables the microrobot to push the object along an arbitrary path. During motion, two conditions are used to ensure the microrobot stays behind the object. These are enforced via a guiding corridor, which the object must remain within. If the object moves to the left of the corridor, the microrobot spins clockwise to reposition it. Conversely, if the object moves to the right, the microrobot spins counterclockwise. The spinning microrobot induces a local microfluidic vortex that subsequently spins the object. This logic forms an autonomous feedback loop that dynamically adjusts the microrobot’s motion to maintain control of the object.  The  pseudo-code is outlined in Algorithm 1.

\section{Performance Metric}
To determine the efficacy of the algorithm, we use completion time and mean absolute error from a desired path as performance metrics.  The desired trajectory $T_d = \{G_1, G_2,\cdots, G_{N_d}\}$ is a circle that consists of $N_d = 100$ nodes or goal coordinates. The completion time is defined as the time it takes to push an object from the starting point $(x_{G_1}, y_{G_1})$ to the end point $(x_{G_{100}}, y_{G_{100}})$. The mean absolute error between the desired path and the objects actual path is defined as $\sum_{n=1} ^{100} \lvert G_i - O_i\lvert /N_d$.

It is important to note the objects actual trajectory $T_a = \{O_1, O_2,\cdots O_{N_a}\}$, consisting of $N_a$ nodes or data points where $O_i = (x_{O_i}, y_{O_i})$, is larger in size than $T_d$.  In other words, $N_a$ is much larger than $N_d$ because the algorithm records the position of the object at each frame. 

As a result, a simple mean absolute error method for measuring the discrepancy between the actual and desired trajectory must be slightly modified to ensure a consistent fair comparison across experiments.  In order to match the size of $T_a$ to the size of $T_d$, the closest point ${O_i}^{closest}=  (x_{O_i}^{closest},y_{O_i}^{closest})$ to each goal position $G_i =(x_{G_i}, y_{G_i})$ in the array is calculated. These points then make up a new $T_a^{new}=\{O_1^{closest},\cdots, O_{N_d}^{closest}\}$ array with size $N_a = N_d = 100$. The mean absolute error is now defined as $\sum_{n=1} ^{100} \lvert G_i - O_i^{closest}\lvert /N_d$. In other words, the distance from each point in $T_a^{new}$ to each point in $T_d$ is calculated, summed, and divided by $N_a = N_d = 100$. This gives a more accurate mean absolute error calculation and allows for a consistent method when comparing different parameters of the algorithm. 

\section{Performance Evaluation}
\subsection{Passive Particle Manipulation}
\begin{figure}[h! tbp]
    \centering
    \includegraphics[width=\linewidth]{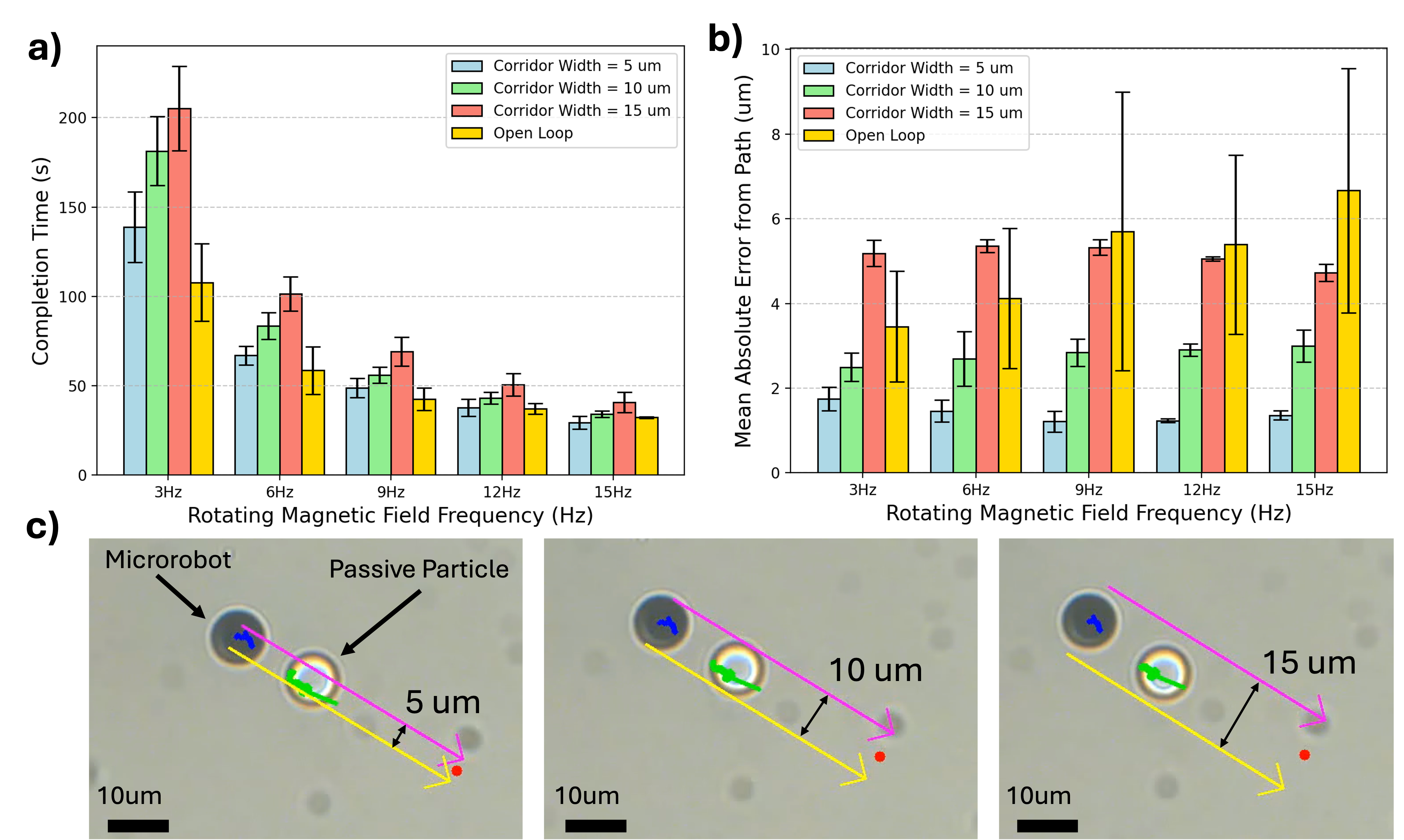}
    \caption{a) Pushing completion times in seconds following a circular trajectory of 100 nodes vs the rotating magnetic field frequency at different corridor widths. b) Mean absolute error ($\mu$m) between the actual path and the desired path vs rotating magnetic field frequency (Hz) and 3 different corridor widths. c) Illustrative bright-field microscopy screenshots of different corridor widths with respect to a 10\,$\mu$m microrobot and a 10\,$\mu$m passive particle.}
    \label{fig:5}
\end{figure}

The algorithms performance was evaluated using a 10\,$\mu$m diameter magnetic rolling microrobot. The microrobot is then tasked to autonomously push a 10\,$\mu$m silica passive microsphere along the circular trajectory consisting of 100 nodes as described above. By summing the distance between each node, we find the total length of the circle to be 538\, $\mu$m. The rotating magnetic field frequency $f$ used for pushing and spinning the object was varied at 3\,Hz, 6\,Hz, 9\,Hz, 12\,Hz, and 15\,Hz.  Additionally, three separate guiding corridor widths were selected to compare the efficacy of the algorithm. The widths were chosen to be one body length of the microrobot ($w = 10\,\mu m$), half a body length ($w = 5\, \mu m$) and one and half times the body length ($w = 15\, \mu m$). See figure \ref{fig:5}c. The approach threshold distance and goal threshold distance were set to 8\,$\mu$m for all experiments. I.e., the object is said to arrive at the goal position once the Euclidean distance between the objects current position and the goal position is less than or equal to 8\,$\mu$m. Finally, the algorithm was also compared to an open loop approach. This open loop approach consisted of manually controlling the microrobots heading angle $\alpha$ using the right joystick of a PS4 gaming controller. 

The bar chart in Figure \ref{fig:5} illustrates the performance results of the algorithm. Four trials for each experiment were conducted, and the bars represent the average of each trial. The error bars represent the standard deviation. Figure \ref{fig:5}a shows completion time results using three separate corridor widths at five different actuation frequencies. It is clear that as the rotating magnetic field frequency increases from 3\,Hz to 15\,Hz, the algorithm is able to push the object along the trajectory at a faster rate. For example, a guiding corridor width of 15\,$\mu$m and an actuation frequency of 3\,Hz results in the algorithm successfully completing the desired trajectory in 205 seconds on average. Alternatively, using the same guiding corridor width of 15\,$\mu$m, and increasing the actuation frequency to 15\,Hz results in a much faster average completion time of 41\,s.  Therefore, it can be concluded that faster actuation frequencies result in faster completion times for the same guiding corridor width. 

Additionally, Figure \ref{fig:5}a indicates shorter guiding corridor widths also decrease the completion time for the same actuation frequency. For example, if we fix the actuation frequency at 3\,Hz, it is clear that a guiding corridor width of 5\,$\mu$m results in the fastest completion time of 139 seconds compared to the 10\,$\mu$m and 15\,$\mu$m widths. Moreover, this trend is observed for all tested frequencies. As a result, we can conclude that shorter guiding corridor widths result in faster completion times.   

\begin{figure}[h!]
    \centering
    \includegraphics[width=\linewidth]{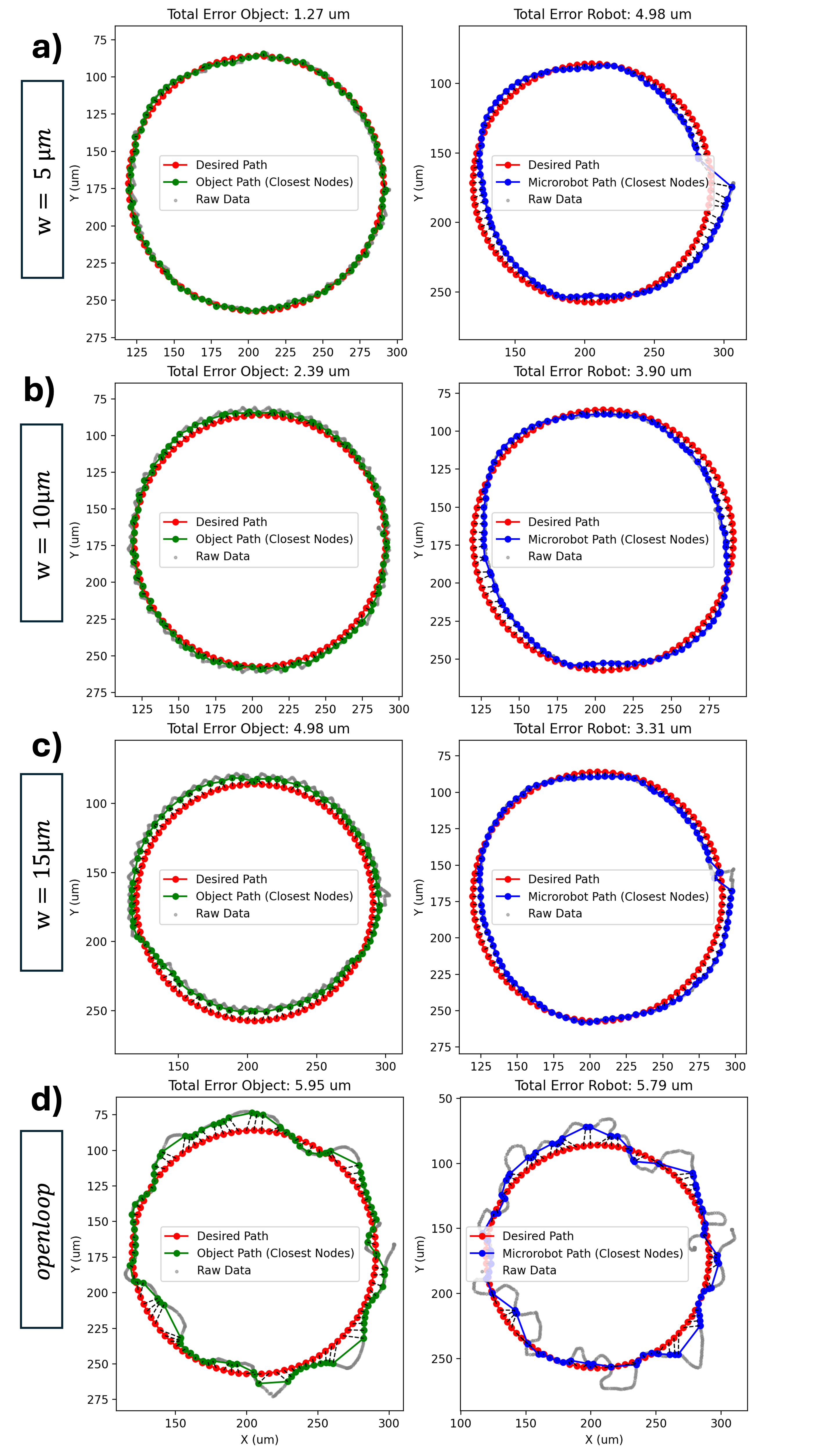}
    \caption{Trajectory plots of an objects trajectory (green) on the left and the associated microrobots trajectory (blue) on the right. The desired trajectory is in red. The raw position data (gray) is also plotted. The rotating magnetic field frequency was set to 12\,Hz. Experiments were conducted at a) 5\,$\mu$m corridor width b) 10\,$\mu$m corridor width, c) 15\,$\mu$m corridor width, and d) manually pushing the object using the open loop method.}
    \label{fig:6}
\end{figure}

It is important to note that manually pushing the object using the open loop method results in the shortest completion time when compared to the algorithm at any guiding corridor width. This is due to the fact that the open loop method does not use the micro vortex readjustment strategy to realign the object back on the path. Therefore, the open loop method can maintain a constant forward speed the entire time. I.e., the algorithm is naturally slower because of the readjustments that the algorithm makes to reposition the microrobot within the corridor. This results in the microrobot and object making less progress and hence longer completion times. 

Figure \ref{fig:5}b shows a plot comparing the mean absolute error between the objects actual path and the desired path at 5 different actuation frequencies. It illustrates how accurately the microrobot is able to push the targeted object along the circular path. Smaller errors correspond with more accurate performance. From the plot, it can be seen that a guiding corridor width of 15\,$\mu$m results in the worst performance across all actuation frequencies.  The average error using a 15\,$\mu$m guiding corridor frequency was found to be approximately 5.1\,$\mu$m across all tested actuation frequencies. 

As we decrease the guiding corridor width, the mean absolute error also decreases.  Therefore, we can conclude that shorter corridor widths result in more accurate performance.  Furthermore, we can conclude the algorithm is more accurate than than the open loop method when using 5\,$\mu$m  and 10\,$\mu$m corridor widths.

From figure \ref{fig:5}, we can conclude which actuation frequency and corridor width resulted in the quickest completion time and smallest mean absolute error.  A guiding corridor width of 5\,$\mu$m and actuation frequency of 15\,Hz resulted in the shortest completion times.  A guiding corridor width of 5\,$\mu$m and actuation frequency of 12\,Hz resulted in the smallest mean absolute error.

\begin{figure}[h! tbp]
    \centering
    \includegraphics[width=\linewidth]{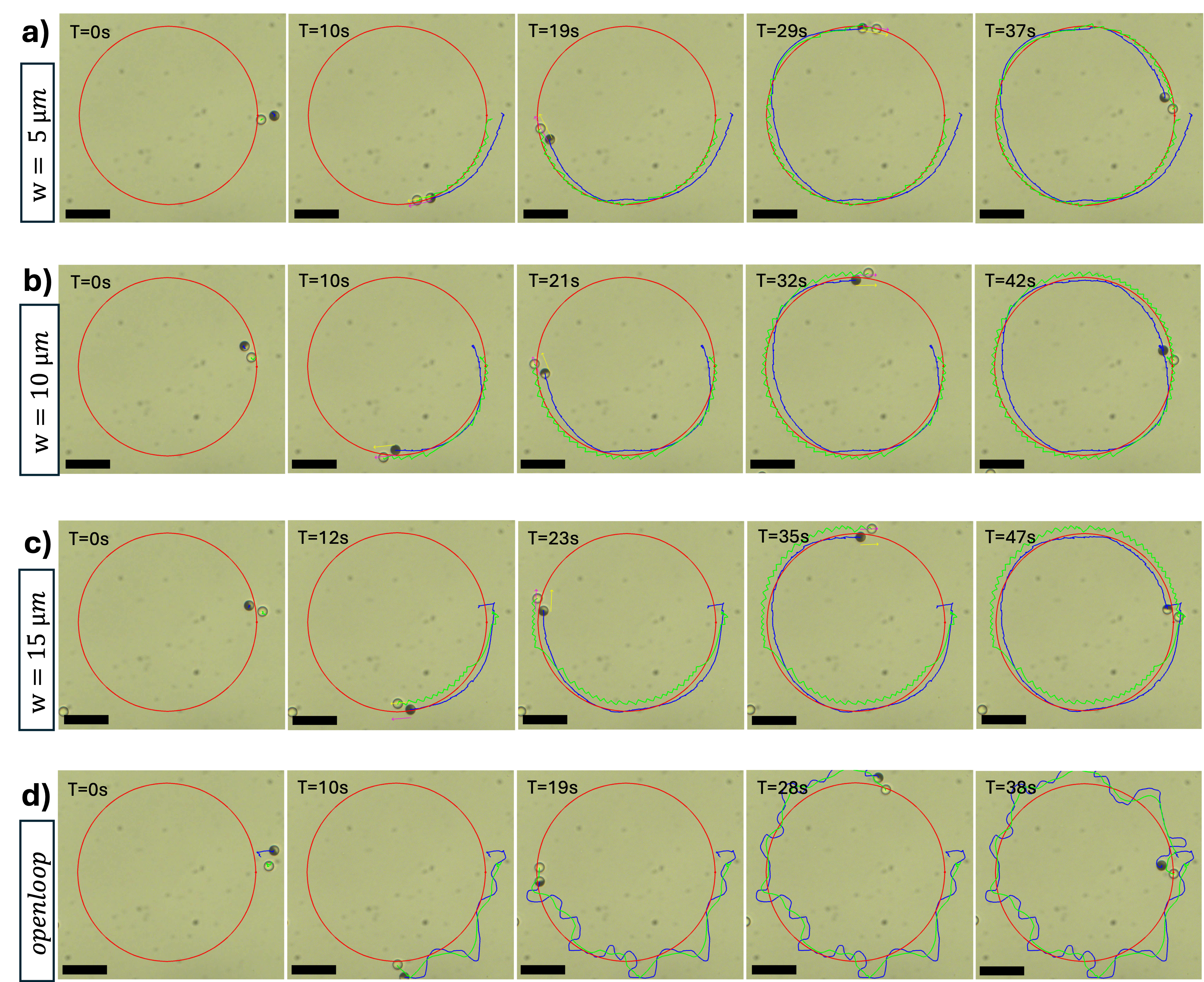}
    \caption{Screenshots of a 10\,$\mu$m rolling magnetic microrobot pushing an object along a circular trajectory at various time instances.  The blue tracks are from the microrobot and the green tracks are the object. The rotating magnetic field frequency is set to 12\,Hz. Scale bar = 40\,$\mu$m. }
    \label{fig:7}
\end{figure}

Figure \ref{fig:6} shows examples of the actual and desired trajectory's for both the object and the microrobot at a 12\,Hz actuation frequency. The red trajectory is the desired trajectory $T_d$ consisting of $N_d = 100$ goal coordinates. The green data points correspond to the actual trajectory of the object. These points are the closest coordinate to each of the desired trajectory coordinate which is $T_a^{new}$ described above. Similarly, the blue data points correspond to the closest coordinates of the microrobot. The black dotted lines display the distances from each coordinate in the desired trajectory to the closest coordinate in each actual trajectory. Finally, the gray data points consist of the actual raw position data measured in real time using the system. 

It is observed from the trajectory plots in Figure \ref{fig:6} that a corridor width of 5\,$\mu$m is most accurate. This is shown by the green coordinates overlapping the red coordinates in Figure \ref{fig:6}a. In this particular experiment, the mean absolute error was measured to be 1.27\,$\mu$m. The error for 10\,$\mu$m (Figure \ref{fig:6}b), 15\,$\mu$m (Figure \ref{fig:6}c), and the open loop method (Figure \ref{fig:6}d) was measured to be 2.39\,$\mu$m, 4.98\,$\mu$m, and 5.29\,$\mu$m respectively. 

In the open loop example from Figure \ref{fig:5}d, it is observed that the microrobot follows a more sporadic trajectory compared to the closed loop examples. As mentioned above, the algorithm does not re-position the microrobot itself to keep the object on the path, but rather repositions the object. This allows for a smoother trajectory of both the object and microrobot as it follows the desired trajectory. 

Finally, Figure \ref{fig:7} show screenshots of a microrobot pushing a passive sphere (See supplemental videos). The rotating magnetic field frequency was set to 12\,Hz.  Figure \ref{fig:7}a, \ref{fig:7}b, \ref{fig:7}c, and \ref{fig:7}d illustrate the 5\,$\mu$m width, 10\,$\mu$m width, 15\,$\mu$m width and open loop experiments respectively. As mentioned above, as we increase the guiding corridor width, the completion time increases. 

Figure \ref{fig:7}d also shows that open loop control resulted in relatively inaccureate trajectories. Guiding the object along the path accurately also becomes more difficult as the rotating frequency is increased. When comparing open loop averages and open loop standard deviations to closed loop, it is clear the algorithm described herein is more accurate and more consistent.

\subsection{Cell Manipulation}
In order to demonstrate biomedical applicability, the algorithm was used to autonomously guide a single cell along a user defined trajectory. We use a guiding corridor width of 5\,$\mu$m and an actuation frequency of 6\,Hz for manipulation.  All cell culture components were from GIBCO, BenchStable, USA. Chinese Hamster Ovary (CHO) cells were maintained in a Dulbecco’s Modified Essential Medium-F12 (DMEM-F12) media supplemented with. 10\% fetal bovine serum (FBS) and 1\% penicillin/streptomycin. Cells were washed with Dulbecco’s phosphate-buffered saline(DPBS) and detached from the culture flask via Tryple Express when they reached the 85\% confluency. Cells were used between passages 3-10.

\begin{figure}[h! tbp]
    \centering
    \includegraphics[width=\linewidth]{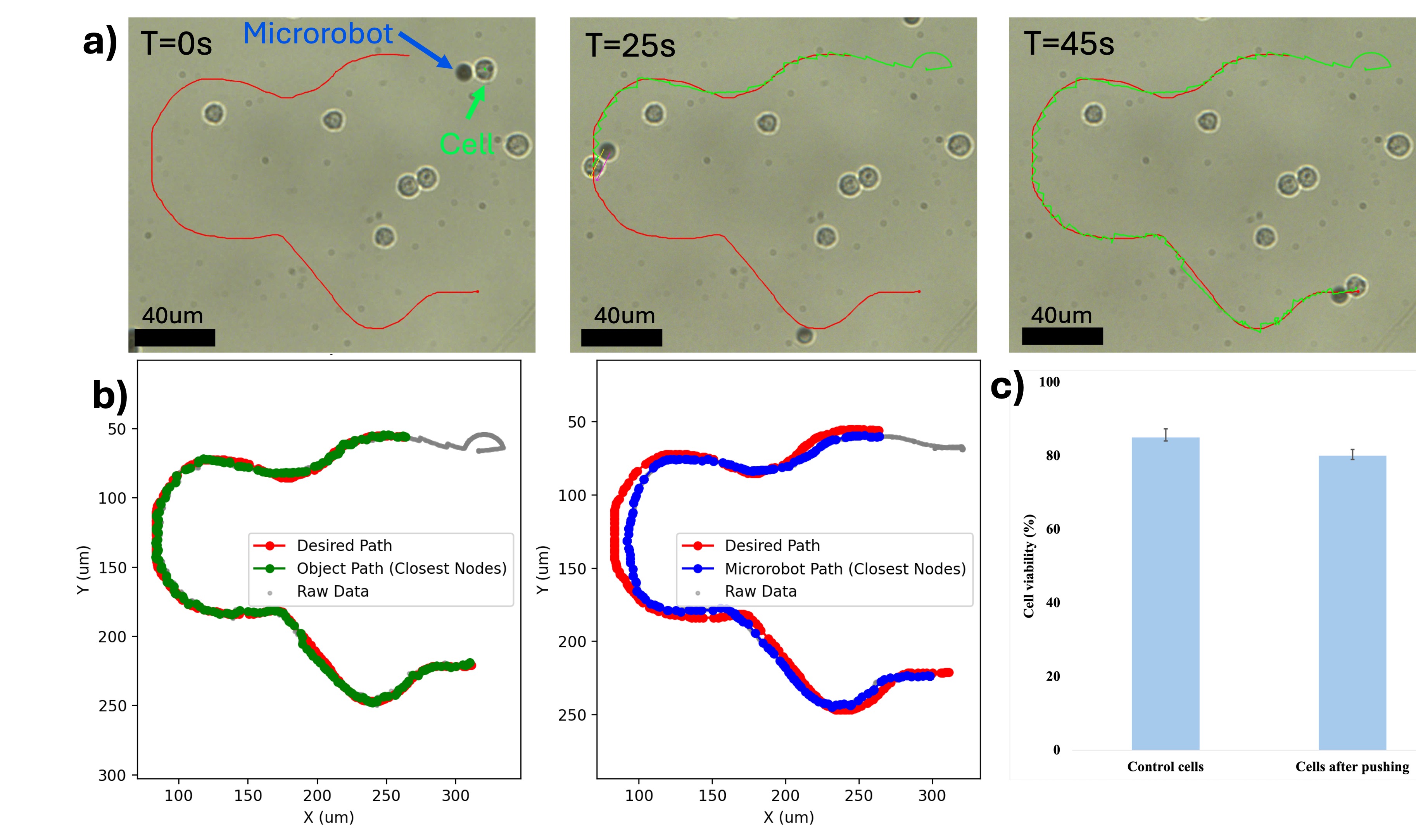}
    \caption{a) Screenshots of algorithm autonomously pushing a CHO cell (green) along an arbitrarily drawn trajectory (red) at various time instances.  The rotating magnetic field frequency is set at 6 Hz. b) The cell and microrobots actual trajectory vs the desired trajectory. c) Cell bio compatibility data. }
    \label{fig:8}
\end{figure}

Figure 8 depicts the results of the experiment. A desired trajectory was drawn with the computer mouse in real time that avoids the other cells in the environment. The path length was measured to be 530\,$\mu$m. The algorithm was successful in autonomously transporting the cell along this trajectory. In this example, the mean absolute error between the cell and the path was 1.35\,$\mu$m and the completion time was approximately 45 seconds. 

Cell viability is a crucial factor for cellular applications using microrobots, thus, we determined the effect of pushing on the CHO cells. Trypan blue cell viability assay was used to evaluate the cell viability after 48 hours. Briefly, cells were incubated under standard cell culture conditions for 48 hours after pushing, and the dye integrates into the cell membrane and differentiates the live and dead cells based on the permeability of the dye in the sample. The viability for non-treated control cells and cells after pushing are 85$\%$ and 80$\%$, respectively (Figure \ref{fig:8}c). Results indicate that the control system and pushing have a negligible impact on the cell viability of the cells which is promising for further cell experiments.

\section{Future Work}
The algorithm suffers from a chattering-like behavior about the guiding corridor edges. In other words, as the microrobot repositions the object back into the corridor, the microrobot may immediately try to push the object back outside the same corridor. This can negatively impact the performance of the algorithm. More chattering is associated with larger guiding corridor widths and therefore slower completion times. This chattering issue can be improved by decreasing the guiding corridor width. Future work make include exploring the minimum guiding corridor width to maximize the completion times of the algorithm. Additionally, creating a more continuous transition between pushing and spinning may also improve the chattering behavior. I.e., instead of a boolean switch from $\gamma = 90^{\circ}$ to $\gamma = 0^{\circ}$ or $180^{\circ}$, there are other angles that combine both the spinning and the pushing mechanisms for smoother manipulation of the object. For example, applying $\gamma = 45^{\circ}$ or $135^{\circ}$ when the object is outside the guiding corridor edges may improve the algorithms performance.

Because the algorithm is data-drive, future experiments could also explore a reinforcement learning implementation.  The inputs being $\alpha$, $\gamma$ and $f$ variables, and the outputs being the mean absolute error from the desired path and the completion time. The reinforcement learning approach could then find optimal combinations of $\alpha$, $\gamma$ and $f$ for which the error and completion times are minimized.

Also, an additional condition can be implemented to ensure an equal distance is maintained between the microrobot and object similar to \cite{10610098}. This is important because there must be a fluid separation layer between the microrobot and object in order for the micro vortex readjustments strategy to work. If the the microrobot and object come to close they can attach resulting in issues. This additional condition will require velocity feedback from the system and may have the potential for improving the robustness and versatility of the algorithm.

Finally, future work will include extending the algorithm to autonomously position multiple cells or objects into patterns. This can be accomplished by selecting multiple objects and goal locations at once and having the microrobot push each object in series. One challenge that remains is to ensure a robust release mechanism so that the objects do not move once they reach their destination.

\section{Conclusion}
This paper presents a novel, vision-based control algorithm for the autonomous manipulation of a micro-object or cell using a magnetic rolling microrobot. The control algorithm's primary method of operation is to push an object within a narrow path toward the target location, readjusting its orientation via spinning, which induces a micro-vortex that rotates the object. This micro-vortex displaces the surrounding fluid, enabling the object to be repositioned. Finally, we demonstrated biomedical applicability by autonomously manipulating a single cell along a trajectory. In conclusion, this paper presents a robust data-driven algorithm for the autonomous manipulation of micro-objects including biological cells.

\bibliographystyle{IEEEtran}
\bibliography{bibliography.bib}

\end{document}


\maketitle

\begin{table}[]
\begin{tabular}{llllllllllllllll}
\multicolumn{7}{l}{path error (um)}                                                           &  & \multicolumn{7}{l}{completion time (s)}                                          &  \\
corridorwidth              & MR Speed (Hz) & Trial 1 & Trial 2 & Trial 3 & Trial 4 & Averages &  & corridorwidth & MR Speed (Hz) & Trial 1 & Trial 2 & Trial 3 & Trial 4 & Averages &  \\
\multirow{5}{*}{5um}       & 3             & 1.98    & 1.97    & 1.39    & 1.66    & 1.75     &  & 5um           & 3             & 144.3   & 151.1   & 150.3   & 109.72  & 138.855  &  \\
                           & 6             & 1.51    & 1.8     & 1.21    & 1.32    & 1.46     &  &               & 6             & 69.6    & 69      & 70.1    & 59.3    & 67       &  \\
                           & 9             & 1.04    & 1.58    & 1.14    & 1.1     & 1.215    &  &               & 9             & 45.9    & 50.5    & 55.6    & 43.6    & 48.9     &  \\
                           & 12            & 1.21    & 1.19    & 1.28    & 1.26    & 1.235    &  &               & 12            & 37.1    & 36      & 44.6    & 33.63   & 37.8325  &  \\
                           & 15            & 1.42    & 1.43    & 1.2     & 1.39    & 1.36     &  &               & 15            & 28.58   & 27.1    & 34.8    & 27.21   & 29.4225  &  \\
\multirow{5}{*}{10um}      & 3             & 2.18    & 2.93    & 2.57    & 2.3     & 2.495    &  & 10um          & 3             & 182.9   & 156.5   & 203.5   & 182.5   & 181.35   &  \\
                           & 6             & 2.29    & 3.58    & 2.73    & 2.16    & 2.69     &  &               & 6             & 82.2    & 80.9    & 94.1    & 76.8    & 83.5     &  \\
                           & 9             & 2.56    & 3.28    & 2.87    & 2.66    & 2.8425   &  &               & 9             & 56.4    & 55.1    & 61.8    & 51      & 56.075   &  \\
                           & 12            & 3.09    & 2.96    & 2.78    & 2.79    & 2.905    &  &               & 12            & 40.5    & 45.3    & 46.5    & 40.3    & 43.15    &  \\
                           & 15            & 2.86    & 2.93    & 3.53    & 2.66    & 2.995    &  &               & 15            & 34.9    & 33.3    & 36.3    & 32.3    & 34.2     &  \\
\multirow{5}{*}{15um}      & 3             & 4.78    & 5.19    & 5.23    & 5.54    & 5.185    &  & 15um          & 3             & 213.5   & 186.3   & 234.7   & 185.7   & 205.05   &  \\
                           & 6             & 5.26    & 5.19    & 5.49    & 5.48    & 5.355    &  &               & 6             & 108.2   & 92.5    & 110.9   & 93.7    & 101.325  &  \\
                           & 9             & 5.2     & 5.54    & 5.15    & 5.4     & 5.3225   &  &               & 9             & 55.9    & 60.4    & 79.2    & 65.8    & 65.325   &  \\
                           & 12            & 5       & 5.1     & 5.01    & 5.1     & 5.0525   &  &               & 12            & 42.9    & 47      & 56.15   & 43.7    & 47.4375  &  \\
                           & 15            & 4.45    & 4.85    & 4.91    & 4.7     & 4.7275   &  &               & 15            & 45.7    & 37.2    & 45.5    & 34.9    & 40.825   &  \\
\multirow{5}{*}{Open Loop} & 3             & 3.73    & 5.18    & 3.2     & 2.61    & 3.68     &  & Open Loop     & 3             & 87.4    & 91      & 140.3   & 132.3   & 112.75   &  \\
                           & 6             & 5.7     & 5.31    & 3.2     & 3.2     & 4.3525   &  &               & 6             & 45.8    & 48.8    & 73.1    & 70.9    & 59.65    &  \\
                           & 9             & 9.1     & 7.94    & 2.54    & 2.7     & 5.57     &  &               & 9             & 37.8    & 37.1    & 46.8    & 51.3    & 43.25    &  \\
                           & 12            & 7.1     & 7.3     & 2.7     & 3.2     & 5.075    &  &               & 12            & 39.3    & 33      & 38.3    & 39.4    & 37.5     &  \\
                           & 15            & 8.67    & 9.4     & 5.34    & 5.2     & 7.1525   &  &               & 15            & 31.9    & 32.7    & 30.2    & 32.6    & 31.85    & 
\end{tabular}
\end{table}

\section{Video Descriptions}
\begin{itemize}

\item Video 1: A 10um magnetic rolling microrobot autonomously pushing a 10um silica passive microsphere using control algorithm. 5 um corridor width at 9Hz rotating magnetic field frequency and 5mT magnetic field strength.

\item Video 2: A 10um magnetic rolling microrobot autonomously pushing a 10um silica passive microsphere using control algorithm. 10 um corridor width at 9Hz rotating magnetic field frequency and 5mT magnetic field strength. 

\item Video 3: A 10um magnetic rolling microrobot autonomously pushing a 10um silica passive microsphere using control algorithm. 15 um corridor width at 9Hz rotating magnetic field frequency and 5mT magnetic field strength.

\item Video 4: A 10um magnetic rolling microrobot manually pushing a 10um silica passive microsphere using gaming controller joystick. 15 um corridor width at 9Hz rotating magnetic field frequency and 5mT magnetic field strength.

\item Video 5: A 10um magnetic rolling microrobot autonomously pushing a single Chinese Hamster Ovarian cell using control algorithm. 15 um corridor width at 9Hz rotating magnetic field frequency and 5mT magnetic field strength.

\end{itemize}